# Thinking on Maps: How Foundation Model Agents Explore, Remember, and Reason Map Environments

Zhiwei WEI [a,b], Yuxing LIU [a], Hua LIAO [a,b], Wenjia XU [c]

[a]*Hunan Normal University, School of Geographic Sciences, Hunan Changsha, China;* [b]*Hunan Key Laboratory of Geospatial Big Data Mining and Application, Hunan Changsha, China;* [c]*School of Information and Communication Engineering, Beijing University of Posts and Telecommunications, Beijing, China.*

Address for correspondence: Wenjia XU. E-mail: xuwenjia@bupt.edu.cn

# Thinking on Maps: How Foundation Model Agents Explore, Remember, and Reason Map Environments

**Abstract:** Map environments provide a fundamental medium for representing spatial structure and supporting navigation, planning, and geographic analysis. Understanding how foundation model (FM) agents acquire, consolidate, and utilize spatial knowledge in such environments is therefore critical for enabling reliable map-based reasoning and applications. However, most existing evaluations of spatial ability in FMs rely on static map inputs or text-based queries, overlooking the interactive and experience-driven nature of spatial understanding. As a result, the mechanisms through which spatial knowledge emerges during exploration and is later leveraged for reasoning remain insufficiently understood. In this paper, we propose an interactive evaluation framework to analyze how FM agents explore, remember, and reason in symbolic map environments. Agents incrementally explore partially observable grid-based maps consisting of roads, intersections, and points of interest (POIs), receiving only local observations at each step. Spatial understanding is then evaluated using a suite of tasks, including direction judgment, distance estimation, proximity judgment, POI density recognition, and path planning. By systematically varying exploration strategies, memory representations, and reasoning schemes across multiple foundation models, we reveal distinct functional roles of these components. Exploration primarily affects experience acquisition but has a limited impact on final reasoning accuracy. In contrast, memory representation plays a central role in consolidating spatial experience, with structured memories particularly sequential and graph-based representations, substantially improving performance on structure-intensive tasks such as path planning. Reasoning schemes further shape how stored spatial knowledge is used, with advanced prompts supporting more effective multi-step inference. Case-based analyses of reasoning traces show that structured memory and reasoning prompts help repair spatial reasoning failures by enabling explicit spatial reconstruction rather than heuristic guessing. We further observe that spatial reasoning performance saturates across model versions and scales beyond a certain capability threshold, indicating that improvements in map-based spatial understanding require mechanisms tailored to spatial representation and reasoning rather than scaling alone. Overall, this work advances experience-driven evaluation of spatial cognition in FM agents and provides insights for designing more reliable map-based reasoning systems.

# 1. Introduction

As foundation models (FMs) and FM agents are increasingly applied in real-world scenarios, such as navigation assistance (Espada et al., 2025) and embodied task planning (Zhai et al., 2025). The need to assess their ability to understand, reason about, and operate within spatial environments has grown rapidly (Xu et al., 2024; Cai et al., 2025). For example, researchers in computer science have begun to explore this challenge in physical or perceptual spaces. They use images, videos, and embodied simulations to test how FMs or FM agents perceive and act in spatially grounded tasks. Several benchmarks have also been developed for this purpose, including VSI-Bench (Yang et al., 2025b), EgoSchema (Mangalam et al., 2023), and OpenEQA (Majumdar et al., 2024). However, geographic or map space is another crucial type of spatial environment. It represents real-world structures into symbolic and organized forms, supporting applications such as geographic analysis, environmental planning, and spatial communication (Ni & Wang, 2024). Therefore, understanding how FMs interpret and reason in such map-based contexts has also become an important emerging topic (Wang et al., 2024; Janowicz et al., 2025).

In response, the GIS community has initiated several efforts to evaluate the ability of spatial understanding in FMs through geographic data or map-based tasks. One major research direction centers on language-based spatial reasoning, where models are tested through tasks such as question answering, geographic knowledge extraction, spatial relation analysis, and code generation (Hu et al., 2023; Mooney et al., 2023; Zhang et al., 2024; Gramacki et al., 2024; Hou et al., 2025; Yang et al., 2025a; Yang et al., 2025c; Xu et al., 2025; Ji et al., 2025). These approaches primarily examine how well models can interpret spatial concepts or complete spatial reasoning tasks through linguistic prompts. However, spatial ability extends beyond linguistic comprehension and also involves visual perception. A second line of research therefore investigates visual and multimodal spatial reasoning. In these studies, models are evaluated on tasks such as chart interpretation, diagram reasoning, and map reading, which integrate spatial and perceptual information (Tao & Xu, 2023; Xu & Tao, 2024; Zhang et al., 2025a; Hong et al., 2025). This stream of research highlights how multimodal inputs can enhance a model's capacity to connect visual context with spatial abstraction. To facilitate systematic comparison, several benchmarks and datasets have also been developed to assess spatial reasoning in geospatial workflows and map-based contexts, such as STBench (Li et al., 2024), GEOBench-VLM (Danish et al., 2025), and GeoAnalystBench (Zhang et al., 2025b).

Despite these promising developments, they still remain limited in capturing dynamic, experience-driven spatial understanding in map environments, and faces two limitations. First, most evaluations are static: they present FMs or FM agents with a complete map or spatial description and require them to answer questions without engaging in any form of spatial interaction. This neglects the process of how a person might explore, observe map space over time (Cen et al., 2024). Second, most settings are out-of-environment: the model is an external reasoner rather than an embedded participant within the spatial context. In contrast, human spatial intelligence arises through active engagement—navigating unknown places, building mental maps, and drawing inferences from experience (De Tinguy et al., 2024; Yang et al., 2025d). Current approaches thus fail to capture the dynamic and embodied nature of spatial cognition in map-based environments.

To bridge this gap, we propose an evaluation framework for assessing how FM agents explore, remember, and reason about symbolic map environments. Instead of being passively presented with a complete map, an agent incrementally explores the environment step by step. At each move, the agent receives only local neighborhood observations and gradually constructs an internal representation of the space, closely resembling how humans explore unfamiliar environments. After exploration, we probe its understanding through tasks including direction judgment, distance estimation, proximity judgment, POI density recognition, and path planning. By systematically varying the agent's exploration strategy, memory representation, and prompting method, we assess under what conditions spatial understanding can emerge in symbolic map settings. Our work advances the study of spatial cognition in FMs by shifting the focus from static map interpretation to interactive, experience-based spatial reasoning. Our contributions are summarized as follows:

- We propose an evaluation framework for assessing spatial understanding in symbolic map environments, where agents explore environments step by step and construct internal representations based on local observations.
- We design a set of probing tasks that cover 15 cities to assess diverse spatial abilities, including direction judgment, distance estimation, proximity judgment, POI density recognition, and path planning.
- We systematically analyze how different exploration strategies, memory representation, and prompting methods influence spatial understanding in foundation model agents.

- Our work bridges the gap between static map interpretation and dynamic, experience-driven spatial cognition, offering insights into how FM agent explore, remember and reason about map space through interaction.

## 2. Related works

### 2.1 Spatial ability evaluation of FMs in physical or perceptual environment

Enabling machines to understand and reason about spatial environments has a long history, which can be traced back to early work in cognitive science and AI on spatial perception and mental mapping (e.g., Kuipers, 1983; Golledge, 1999). With the emergence of FMs and their increasing deployment in real-world applications, the spatial ability of machines has received renewed attention. In this context, we focus on the spatial ability evaluation of FMs, which can broadly be categorized into three parts: image-based, video-based, and embodied spatial reasoning.

The image-based evaluation primarily tests FMs on tasks such as object localization, spatial relation classification, or distance/depth estimation. For example, the benchmarks such as SpatialVLM (Chen et al., 2024), MM-Spatial (Daxberger et al., 2025) introduce 3D spatial VQA datasets to evaluate metric spatial reasoning from 2D input. These studies reveal that while FMs demonstrate strong performance in relative-position recognition, they still struggle with metric and geometric reasoning, such as occlusion handling. Even models pre-trained with large 3D datasets struggle to generalize to unseen spatial configurations (Chen et al., 2024). Zhou et al. (2024) further observed that MLLMs lack robust viewpoint invariance and fail to generalize spatial relationships across unseen perspectives. Similar analyses like SpatialRGPT further reveal that these models rely heavily on statistical co-occurrence rather than genuine geometric reasoning (Cheng et al., 2024).

The second strand moves beyond static imagery to video-based benchmarks, which add temporal dynamics and causal dependencies. Benchmarks such as VSI-Bench (Yang et al., 2025b), EgoSchema (Mangalam et al., 2023), and OpenEQA (Majumdar et al., 2024) evaluate the capacity of FMs to infer spatial relations, track entities, and plan routes from egocentric video streams. These tasks require models to process sequential frames, infer distance and direction, and maintain spatial consistency across time. For example, VSI-Bench tests MLLMs on object counting, spatial direction estimation, and route inference in dynamic indoor scenes. These evaluations reveal that FMs can encode short-range motion patterns but often fail to form a global spatial map that persists over time, leading to instabilities in route reasoning. Recent surveys also show that although models such as GPT-4V and Gemini exhibit temporal fusion capabilities, they still lack long-term spatial coherence and dynamic memory (Zheng et al., 2025). Thus, the latest MLLM benchmarks emphasize that integrating temporal grounding, memory modules, and process-level reward signals (e.g., Spatial-R1, Video-R1) is critical to achieving consistent spatial understanding in video-based contexts (Feng et al., 2025).

The third strand of work focuses on embodied and interactive environments, where FMs perceive, act, and reason within simulated or physical space. Recent work emphasizes the interplay of perception, memory, and action (An et al., 2025). Systems such as VL-Nav (Du et al., 2025) integrate multi-modal pretraining with episodic memory modules to support path planning and object manipulation under partial observability. Likewise, frameworks like SpatialAgent (Wu et al., 2025) demonstrate that coupling perception and memory enables higher-level spatial abstraction and more robust environmental reasoning. More recently, the NaviTrace benchmark (Windecker et al., 2025) provides a standardized evaluation of embodied navigation capabilities of vision-language models by having models predict 2D navigation traces from real-world images and instructions. Evidence across these frameworks suggests that coupling a vision model with episodic memory or navigation modules leads to stronger spatial abstraction capabilities.

In summary, the growing literature on image-based, video-based, and embodied spatial reasoning has greatly advanced our understanding of how FMs process and utilize spatial information. However, symbolic spaces such as maps also constitute a fundamental mode through which humans understand, communicate, and reason about the world. These abstract, structured, and representation-driven environments differ substantially from perceptual or egocentric ones, yet they have received far less attention in current FM research.

## 2.2 Spatial ability evaluation of FMs in map-based environments

Spatial ability evaluations in geographic contexts are usually large-scale, symbolic, and context-dependent. Montello (1998) identified four levels of spatial cognition, including figural, vista, environmental, and geographic, where the latter two rely on integrating information gathered across extensive movement or symbolic representations such as maps. With the rise of FMs, the GIS community has also begun to examine whether these models can exhibit similar symbolic and large-scale reasoning abilities within map-based environments.

The researchers first explore how LLMs handle geographic or spatial reasoning tasks using natural language. These works typically involve spatial question answering, geographic knowledge extraction, spatial relation classification, and geospatial code generation. Mooney et al. (2023) evaluated ChatGPT's geospatial reasoning within GIS workflows and observed that while it could generate plausible explanations, it often neglected real-world spatial constraints. Hu et al. (2023) found that incorporating explicit geographic cues into disaster-related messages improved model understanding of spatial relations. Li et al. (2024) developed STBench, a benchmark containing 16 spatial-temporal reasoning tasks, and identified that while LLMs perform well on simple categorical relations, they struggle with topological and geometric reasoning. Subsequent studies (Ji et al., 2025; Yang et al., 2025a) highlighted that prompt engineering and contextual augmentation can improve performance on spatially grounded tasks. Related work also focuses on code-based spatial reasoning, where LLMs are prompted to solve geospatial problems using Python, SQL, or GIS scripting. Xu et al. (2024) compared GPT-4 and CodeLLaMA across 61 geospatial coding problems and found a persistent gap between general-purpose models and domain expectations. To narrow this gap, the GeoCode-GPT framework introduced structured prompts and task abstraction modules that align better with geospatial reasoning needs (Hou et al., 2025).

Beyond text-based tasks, recent research extends spatial reasoning evaluation to multimodal contexts by introducing visual or map-based inputs. Tao and Xu (2023) and Xu and Tao (2024) analyzed GPT-4V's ability to interpret choropleth maps and infographics, showing that the model often relied on superficial visual cues without grasping the underlying spatial logic. Danish et al. (2025) proposed GEOBench-VLM, a comprehensive benchmark for evaluating visual–language models such as Gemini, Claude, and GPT-4V across tasks including land-use detection, spatial alignment, and symbol decoding, revealing substantial inter-model variation. Huang et al. (2025) presented VisFactor, which dissects visual cognition into fundamental primitives—color, shape, and position—and demonstrated that spatial consistency and object localization remain key limitations for current multimodal models. These findings collectively suggest that while multimodal foundation models can integrate visual and spatial information, they still fall short in developing coherent spatial abstractions across map scales or layers.

To address these limitations, several studies have proposed hybrid or domain-adaptive architectures that integrate language-based reasoning with structured spatial representations. Yang et al. (2025a) introduced HybridMind, which transforms qualitative textual reasoning from LLMs into structured spatial data and connects them with GIS tool chains for downstream analysis. Xu et al. (2025) developed RS-Agent, a modular system combining LLMs with remote sensing models and spatial knowledge bases, achieving improved performance in complex geospatial workflows such as change detection and map classification. In the cartographic domain, frameworks like MapReader (Zhang et al., 2025a) and MapColorAI (Yang et al., 2025c) emphasize the importance of visual–symbolic association learning, demonstrating that pretraining on map-specific datasets can enhance model accuracy in layout, color, and symbol understanding.

In summary, these studies have made substantial progress in advancing the spatial understanding and reasoning capabilities of foundation models, providing valuable insights, benchmarks, and hybrid frameworks for geospatial intelligence. However, most of these evaluations remain static and disembodied, which mainly test model reasoning on fixed maps or descriptions rather than through active interaction. In contrast, human geographic reasoning emerges from continuous engagement with maps, exploring and remembering spatial layouts, and constructing internal mental representations through experience. Researchers in the computer vision and robotics communities have also begun to explore interactive spatial reasoning in physical environments, where agents learn through perception, memory, and embodied action (as discussed in *Section* **2.1**). Therefore, there is a need to move beyond passive map interpretation toward interactive and exploratory frameworks, enabling FM agents to build and update internal representations dynamically and achieve more human-like spatial understanding.

## 3. Problem formulation

We formulate map-based spatial cognition as an interactive process inspired by how humans explore unfamiliar environments, in which an agent incrementally observes, acts, and accumulates spatial experience before reasoning about the environment. Under this formulation, spatial understanding does not emerge from a single input–output query, but from the interaction between observation, action, memory, and reasoning under partial observability. The overall framework is illustrated in Figure 1, and this section is organized into three components that align with it. First, we define the map environment (*Section* **3.1**), specifying how spatial elements such as roads, intersections, and POIs are represented. Second, we describe the agent design (*Section* **3.2**), including how the agent perceives local neighborhoods, moves within the map, and adopts different exploration strategies and memory representations. These design choices jointly determine how spatial experience is acquired, stored, and utilized during inference. Third, we introduce a set of spatial understanding evaluation tasks (*Section* **3.3**), which probe different aspects of spatial cognition through question answering under different reasoning schemes.

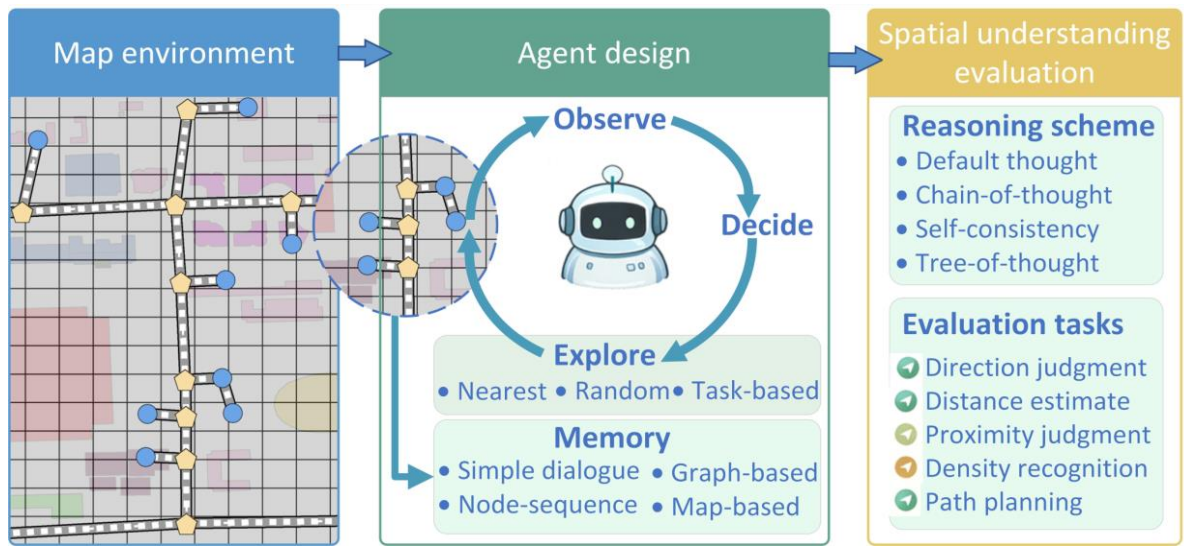

Figure 1. Framework

## 3.1 Map environment

Since our goal is to evaluate how agents explore, remember, and understand map environments, we adopt a grid-based road and POI representation that reflects the most common form of map navigation (Guo et al., 2018). The agent explores the environment step by step, similar to how users traverse maps in practice. Based on this design, we construct a symbolic grid world that integrates geometric, topological, and semantic information. Thus, we describe the environment setup, including the (1) map representation, (2) data source, and construction.

### 3.1.1 *Map representation*

The map environment is represented as a discrete grid structure that encodes both roads and points of interest (POIs). As shown in Figure 2, each grid cell is assigned a categorical symbol denoting its spatial type: road, road intersection, or POI. Unique identifiers are assigned to every intersection and POI. To ensure meaningful navigation, each POI is linked to its nearest road cell, and connectivity between POIs is defined strictly through road paths. This guarantees that moving from one POI to another always occurs along the underlying road network. The grid thus provides a simplified yet interpretable foundation for spatial reasoning, where roads and intersections define the navigable structure, and POIs serve as semantic anchors that support spatial reference and memory formation.

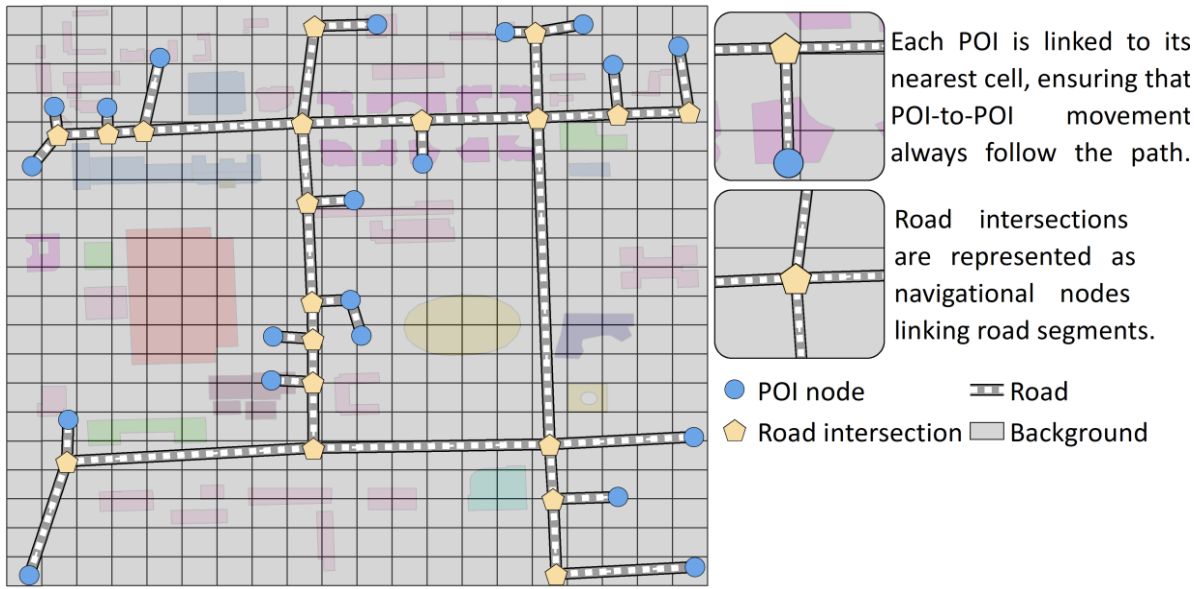

Figure 2. Map representation

### 3.1.2 *Data source and map construction*

To ensure that the experimental environment reflects diverse real-world urban patterns, we collected small-area map data from 15 cities across China, the United States of America (USA), and Europe. Each sample covers a representative urban block containing both road networks and points of interest (POIs) that characterize typical functional and morphological structures of cities in different cultural and planning contexts. The datasets were obtained from OpenStreetMap (OSM), from which the road and POI vector layers were extracted and rasterized into symbolic grids. Each grid cell corresponds to a fixed spatial resolution unit and encodes the presence of a specific spatial feature (road, road intersection, POI, or background). All maps are normalized to a fixed dimension of 20 × 20 cells to maintain comparable spatial complexity across samples. Within each grid, 9 to 21 POIs are retained according to their actual density patterns in the corresponding city block. The details of the dataset are summarized in Table 1.

Table 1. Overview of the symbolic map datasets used in the experiments, showing the distribution of POIs, road intersections, and main roads for each selected city. All maps are normalized to 20 × 20 grids; auxiliary connecting roads linking POIs are not included in the reported road counts.

| Area | City name | POI Count | Road intersection Count | Main Road Count |
|---|---|---|---|---|
| China | Beijing | 21 | 22 | 4 |
|  | Shanghai | 9 | 10 | 4 |
|  | Guangzhou | 17 | 13 | 9 |
|  | Changsha | 12 | 19 | 4 |
|  | Wuhan | 20 | 23 | 10 |
| The USA | New York | 15 | 26 | 7 |
|  | Los Angles | 15 | 17 | 6 |
|  | San Francisco | 15 | 30 | 12 |

|  | Chicago | 15 | 26 | 6 |
|---|---|---|---|---|
|  | Toronto | 15 | 23 | 6 |
|  | London | 15 | 25 | 8 |
|  | Paris | 15 | 17 | 9 |
| Europe | Rome | 15 | 16 | 6 |
|  | Berlin | 15 | 23 | 9 |
|  | Vienna | 15 | 24 | 15 |
|  | Average | 15.27 | 20.93 | 7.67 |

## 3.2 Agent design

This section describes the agent design illustrated in Figure 1, including (1) the observation and movement mechanisms that govern agent-environment interaction, (2) exploration strategies that determine spatial experience acquisition, and (3) memory representations that consolidate experience for later reasoning.

### 3.2.1 *Observation and movement*

To simulate realistic map-based exploration, the interaction between the agent and the environment is designed around two complementary components: observation and movement. Observation defines how much of the surrounding environment the agent can perceive at any given moment, while movement determines how the agent acts and updates its position within the grid. The observation and movement are shown in Figure 3.

**(1) Observation**

The environment is defined as partially observable: at each time step, the agent can only perceive a local neighborhood around its current position within a fixed view range (e.g., a 5 × 5 window), as shown in the left part of Figure 3. This design simulates the limited field of view experienced during human map reading and exploration. The agent receives structured symbolic information within this local view, including the relative positions of nearby POIs and road intersections. Global information about the map layout is not directly accessible; instead, the agent must integrate local observations over time to construct an internal representation of the overall environment.

**(2) Movement**

Movement in our setting consists of two components: the action space, which defines how the agent can move within the grid, and the movement mode, which determines how the agent moves to its destination during exploration.

**Action space**: the agent operates within a discrete eight-directional action space, corresponding to the Moore neighborhood in grid-based environments. At each step, it can move to any of the eight adjacent cells—up, down, left, right, or diagonally.

**Movement mode**: in grid-based navigation settings, an agent move can generally be designed in two ways: one fixes the target while varying the navigation policy, allowing the agent to try different path-finding behaviors to reach the same destination; another fixes the navigation policy while varying the target selection, where the path is deterministic but the exploration order changes (Hart & Moore, 1973). Our design follows the second paradigm. Agent move is structured as goal-driven navigation, in which the exploration strategy (see ***Section* 3.2.2**) determines which POI to visit next, while the movement between POIs always follows the shortest available road path, as shown in the right part of Figure 3. This design simulates high-level map-based exploration—similar to how a user selects the next destination on a map while relying on a navigation system to compute an efficient route. As shown in Figure 3, once an agent arrives at a POI, it selects its next destination from the currently observable neighborhood using an exploration strategy. If an unvisited POI is chosen, the agent follows the shortest path (highlighted in green) using the predefined movement actions.

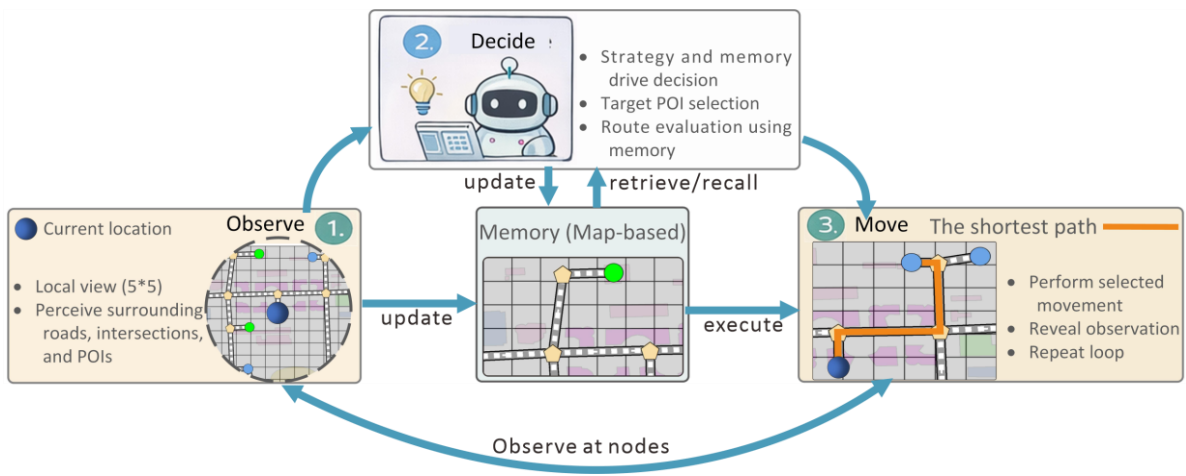

Figure 3. Observation and movement

### 3.2.2 *Exploration strategies*

Under the goal-driven movement described in ***Section* 3.2.1**, the agent's movement is determined by a two-step process: first selecting a target POI according to an exploration strategy, and then navigating to that target along the shortest available road path. By varying the target-selection strategy while keeping the routing mechanism fixed, we isolate the influence of exploration behavior on spatial learning. To ensure that the agent sufficiently covers the environment, each POI must be visited at least *n* times across the exploration process, regardless of strategy. Three strategies are applied in this evaluation, as follows.

(1) Nearest-POI Strategy (NPS)

At each exploration step, the agent selects the nearest less-explored POI within its observation window. This greedy, locally expanding strategy reflects how humans often explore unfamiliar environments by gradually extending outward from known regions.

(2) Random-Visible Strategy (RVS)

The agent randomly chooses a POI from those currently visible within its observation window. This introduces controlled stochasticity while grounding decisions in what the agent can locally perceive. It simulates exploratory behaviors such as opportunistic detours or curiosity-driven choices, allowing us to examine how irregular or non-systematic exploration affects the formation of spatial knowledge.

(3) Task-Driven Strategy (TDS)

The agent follows a predefined sequence of POI pairs, navigating between them regardless of spatial proximity. This strategy models task-driven scenarios—for example, when external instructions dictate exploration order, such as following a checklist or executing a user-defined route. It enables us to analyze whether structured but non-spatial exploration sequences can still support coherent internal map representations.

For the NPS and RVS, the observation window may occasionally contain no eligible POIs. In such cases, the agent selects a new starting POI globally and resumes exploration from that point, preventing stagnation due to partial observability. Across all strategies, exploration continues until every POI has been visited at least $n$ times, ensuring balanced spatial exposure and enabling meaningful evaluation of memory and reasoning performance. From a behavioral perspective, NPS and TDS correspond to structured exploration strategies, as target selection follows explicit spatial or task rules, whereas RVS represents a stochastic exploration strategy that introduces randomness under local observability

### 3.2.3 *Memory representation*

As the agent explores the symbolic map environment, it continuously receives local observations of road intersections and POIs via the perception–action loop. The memory mechanism determines how these sequential observations are encoded, stored, and later retrieved to support spatial reasoning. We examine four distinct memory representations as shown in Figure 4.

**(1) Simple Dialogue Memory (SDM)**

Simple memory represents the agent's default and unstructured form of memory, which naturally arises from the dialogue flow during interaction. Under this setting, no explicit memory structure or spatial organization is imposed. Instead, the agent relies solely on the accumulated conversational context while using FM, in which local observations, past actions, and intermediate responses are implicitly recorded as text tokens in the dialogue history.

**(2) Node-Sequence Memory (NSM)**

Node-sequence memory is designed to mimic the way humans sometimes rely on rote, sequential recall—remembering what was seen at each place and the order in which places were visited, without forming any structured or map-like representation of the environment. Under this mechanism, the agent simply records a chronological sequence of salient nodes encountered during exploration, namely POIs and road intersections. At each node, the memory stores all locally visible information—such as connected road segments, nearby intersections, and POIs—reflecting a low-level, egocentric accumulation of observations. The memory also logs the exact route taken between POI pairs, preserving the ordered list of intersections traversed along each path. To prevent unnecessary duplication during repeated exploration, the agent checks the memory each time it arrives at a POI or intersection. If the node or the POI-to-POI route has already been stored, it is not added again.

**(3) Graph Memory (GM)**

Classic studies in spatial cognition show that humans often represent environments as networks of landmarks and connecting routes, rather than as simple sequences of visited places (Kumaran, 2005; Garvert et al., 2017). Drawing on this insight, we design the graph memory to model the environment as a relational structure, which comprises nodes (POIs and intersections) and edges (road segments). Instead of storing observations as a chronological list, the agent incrementally builds a topological graph as exploration proceeds. At each step of exploration, the agent updates the graph based on its new local observation and the route taken to the current node. When the agent arrives at a POI or intersection, it checks whether this node already exists in memory.

- If it does not exist, the node is added along with its locally visible connectivity (e.g., which intersections or POIs are connected via outgoing road segments).
- If it already exists, the agent updates the node's adjacency information only if newly observed connections were not previously recorded.

The same applies to edges: whenever the agent travels along a road segment between two nodes, the corresponding edge is added if missing. Over time, this produces a connectivity-preserving representation of the environment.

**(4) Map Memory (MM)**

Human spatial cognition research also suggests that people can form map-like representations of environments that preserve global structure rather than only local connectivity (Tolman, 1948; Ekstrom & Hill, 2023). This mechanism has also been proved and applied by Yang et al. (2025b) for the spatial understanding of MLLM. It explicitly stores scene-level spatial structure for later recall and reasoning. In our implementation, map memory aggregates observations across exploration to approximate a global spatial map that preserves both the geometry and the semantic layout of the environment. Each time the agent arrives at a POI or intersection, it updates this map with the following information:

- the estimated spatial position of the node,
- the road geometry connecting it to neighboring nodes,
- and the semantic attributes associated with the POI or intersection.

When a node is visited for the first time, it is added to the map at a position inferred from the agent's local observation. If the node already exists, its position or spatial relations are refined using new evidence. Similarly, road segments are inserted into the global map as continuous spatial curves rather than simple graph edges, enabling the agent to reconstruct not only connectivity but also layout shape. Over repeated exploration, these updates gradually yield a coherent 2D representation of the environment—essentially a simplified but globalized map.

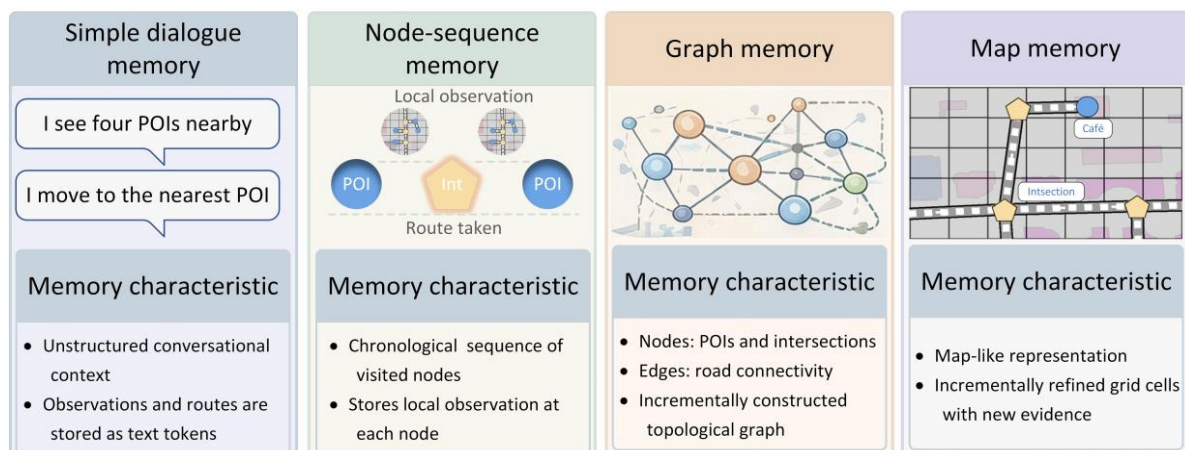

Figure 4. Memory

## 3.3 Spatial understanding evaluation

### 3.3.1 *Reasoning schemes*

In addition to exploration and memory strategies, the prompting strategy used when querying the agent can also influence its spatial reasoning performance. To ensure a fair and comprehensive evaluation, we adopt four widely used reasoning prompts in our task setup.

- **Default Thought (DT)**. In this setting, the model is instructed to directly produce an answer to the given spatial task without any additional reasoning instructions.

- **Chain-of-Thought (CoT)**. In this setting, we append a short instruction 'Let's think step by step' to the end of each task prompt.
- **Self-Consistency with CoT (SC-COT)**. Following Wang et al. (2022), we sample multiple reasoning chains by setting the model temperature to 1.0 and generating several CoT answers (five in our experiments). The final prediction is chosen by majority vote over these sampled responses.
- **Tree-of-Thoughts (ToT)**. In this setting, we divide the model's reasoning into two stages. In the first stage, the model is asked to propose and select a tentative reasoning plan for solving the spatial task (e.g., which landmarks to recall, which routes or relations to consider). In the second stage, conditioned on the chosen plan, the model generates several candidate answers and then selects the one it judges most plausible as the final prediction.

### 3.3.2 Evaluation tasks

To evaluate how different exploration strategies and memory representations support spatial understanding, we design a standardized set of map-based cognitive tasks for each city. Human spatial cognition in unfamiliar environments relies on several fundamental abilities—such as orienting oneself, estimating distances, judging proximity, recognizing local spatial density, and reasoning about optimal routes (Han et al., 2025). These abilities are also essential for effective map reading and navigation. Therefore, our task set is designed to capture these core dimensions in a simplified but principled manner.

All tasks are generated directly from the POI and road-intersection information contained in each symbolic grid map. This ensures that performance differences can be attributed to the agent's exploration and memory mechanisms, rather than to dataset bias or task-specific priors. Furthermore, to enable cross-city comparison, each city block produces the same five categories of spatial tasks, with comparable cognitive demands and difficulty levels. The detailed tasks are shown in Table 2.

Table 2. Evaluation tasks.

| Category | Description | Example | Count & Design Rationale |
|---|---|---|---|
| Direction judgement (DJ) | Determine the relative direction of one POI from another. | 'What is the direction of POI B relative to POI A?' (A. North B. South C. East D. West) | **8 items per city**: Organized as four symmetric pairs, for each pair, one item asks the direction from X to Y, and another asks the reverse direction Y to X. |
| Distance estimation (DS) | Estimate the distance between two POIs | 'The distance from POI A to POI B is closest to which interval?' (A. Range 1 B. | **4 items per city**: Each item selects a POI pair at different true distances (short/long) and |

| | | Range 2 C. Range 3 D. Range 4) | requires matching to predefined distance intervals. |
|---|---|---|---|
| Proximity judgement (PJ) | Choose which POI is closest to a reference POI. | 'Which POI is closest to A?' (A. POI 1 B. POI 2 C. POI 3 D. POI 4) | **4 items per city**: Generated by selecting 1 reference POI and 4 randomly sampled alternatives. |
| POI density recognition (PDR) | Identify which region has the highest (or lowest) POI density. | 'Which area has the highest POI density?' (A. Region 1 B. Region 2 C. Region 3 D. Region 4) | **4 items per city:** Two items select the highest-density region; the other two select the lowest-density region. |
| Path planning (PP) | Select the shortest path among several candidate routes. | 'Which route from A to D is shortest?' (A. Route 1 B. Route 2 C. Route 3 D. Route 3) | **4 paths per city**: Each city yields two long-distance route tasks (global routing) and two short-distance route tasks (local routing). |

# 4. Evaluation

To systematically examine how spatial understanding emerges in map environments, our evaluation is organized into three consecutive phases. Phase I investigates the effect of different exploration strategies on spatial experience acquisition, while keeping memory and reasoning fixed. Phase II focuses on memory representations, analyzing how different memory structures consolidate accumulated experience and influence spatial reasoning performance. Phase III evaluates prompting and reasoning schemes, examining how stored spatial knowledge is utilized during inference.

## 4.1 Phase I: effect of exploration strategies

### 4.1.1 *Experimental setup*

Phase I aims to isolate the effect of exploration strategies on spatial understanding. To this end, we fix all other components of the agent and vary only the exploration strategy during the exploration phase. Specifically, all agents operate in the same symbolic map environments described in *Section* **3.1**. Each environment is a 20 × 20 grid constructed from real-world urban road networks and POIs, and is partially observable with a fixed local view window of 5 × 5 cells. The action space is identical across all conditions, allowing movement to the eight neighboring cells at each step. To avoid confounding effects introduced by navigation policies, the movement between two POIs is always computed as the shortest available road path. Under this design, exploration strategies differ only in how the next target POI is selected, while the underlying routing mechanism remains fixed. During **Phase I**, we adopt a minimal cognitive configuration. The agent uses the simple dialogue memory described in *Section* **3.2.3**, which relies solely on the dialogue context without any explicit spatial structure. In addition, we apply the simple thought prompting scheme described in ***Section* 3.3.1**, in which the model directly outputs an answer without any additional reasoning instructions. The agent's spatial understanding is evaluated using the standardized set of map-based tasks described in ***Section* 3.3.2**. Performance is measured by task accuracy, reported both overall and by task category.

Three exploration strategies are compared: Nearest-POI Strategy (NPS), Random-Visible Strategy (RVS), and Task-Driven Strategy (TDS), as defined in ***Section* 3.2.2**. For each strategy, exploration continues until every POI in the map has been visited at least 1 time, ensuring comparable spatial exposure across strategies.

### 4.1.2 Result

Table 3 summarizes the spatial reasoning performance of different foundation model agents under NPS, RVS, and TDS while keeping the memory representation and prompting scheme fixed. We analyze the results from three aspects: tasks, exploration strategies, and models.

**(1) Task-type sensitivity**

Across all models and exploration strategies, a clear hierarchy emerges among task categories. Direction judgment (DJ) and proximity judgment (PJ) generally achieve relatively higher accuracy across models. For example, Gemini-2.5-Pro maintains high performance on DJ (91.67–94.17%) and PJ (90.00–95.00%) under all three strategies. Even for the weakest cases, DJ accuracy remains around 46.67% (e.g., Qwen-3 under TDS), and PJ does not drop below 51.67% (e.g., GPT-5.2 under NPS).

In contrast, POI density recognition (PDR) remains the most challenging task. Across models and strategies, PDR accuracy is consistently lower than DJ and PJ, often staying below 50%, and in some cases below 30% (e.g., DeepSeek-V3.2: 18.33-31.67%), indicating persistent difficulty in global density estimation.

The remaining tasks, distance estimation (DS) and path planning (PP), exhibit larger variability across models. For example, Gemini-2.5-Pro achieves very high DS accuracy (93.33–95.00%) and moderate PP (61.67–71.67%), whereas GPT-5.2 shows much lower DS performance (20.00–31.67%) and PP ranging from 46.67% to 55.00% across strategies. This contrast indicates that DS and PP are more dependent on model capability than DJ and PJ.

Overall, the results show that DJ and PJ remain consistently high-performing tasks even in the lowest-performing cases, PDR is consistently low-performing, while DS and PP display pronounced cross-model variation. This task-dependent pattern is consistent across exploration strategies, suggesting that task characteristics dominate performance variance at this stage, while exploration primarily modulates performance within task-specific bounds.

**(2) Effect of exploration strategies**

Comparing exploration strategies, differences in overall accuracy are generally modest, while task-level effects are more apparent. For most models, total accuracy varies within a narrow range across strategies. For example, GPT-5.2 ranges from 43.89% (NPS) to 45.83% (RVS) and 44.45% (TDS), and Gemini-2.5-Pro remains nearly unchanged (79.17-79.44% across all three strategies).

At the task level, RVS tends to slightly improve PP performance for several models. Gemini-2.5-Pro achieves 71.67% PP under RVS, compared with 61.67% (NPS) and 68.33% (TDS), and GPT-5.2 improves from 50.00% (NPS) to 55.00% (RVS). In contrast, TDS shows relatively stronger PJ performance for some models, such as Gemini-2.5-Pro (95.00%) and DeepSeek-V3.2 (63.33%), compared with NPS and RVS. However, PDR remains consistently low regardless of strategy, indicating limited benefits of strategy choice for global density reasoning.

Overall, the results suggest that while exploration strategy can influence specific tasks, its impact on overall performance is limited when memory and prompting are fixed

**(3) Cross-model consistency**

Across all three exploration strategies, model capability dominates performance differences. Gemini-2.5-Pro consistently achieves the highest total accuracy, remaining around 79% under NPS, RVS, and TDS, with strong and stable performance across all task categories. Claude-4.5 and Qwen-3 form a middle tier, with total accuracy typically in the 53–71% range depending on strategy. In contrast, GPT-5.2 and DeepSeek-V3.2 exhibit substantially lower overall accuracy, generally below 50%, and show greater sensitivity to exploration strategy in DJ and DS. For example, DeepSeek-V3.2's DS varies from 56.67% (NPS) to 45.00% (TDS), while its PDR remains low across all strategies.

These patterns indicate that differences between models are larger than differences between exploration strategies, and that exploration strategy alone cannot compensate for limitations in model-level spatial reasoning capability.

Table 3. The spatial reasoning performance of different foundation model agents under three exploration strategies (NPS, RVS, and TDS).

|  | Category | gpt-5.2 | gemini-2.5-pro | claude-4.5 | qwen-3 | DeepSeek-V3.2 |
|---|---|---|---|---|---|---|
| NPS | DJ | 53.33% | 93.33% | 79.17% | 52.50% | 51.67% |
|  | DS | 20.00% | 95.00% | 56.67% | 65.50% | 56.67% |
|  | PJ | 51.67% | 91.67% | 81.67% | 73.33% | 60.00% |
|  | PDR | 35.00% | 41.67% | 48.33% | 50.00% | 18.33% |
|  | PP | 50.00% | 61.67% | 58.33% | 73.33% | 23.33% |
|  | Total | 43.89% | 79.44% | 67.22% | 61.11% | 43.61% |
| RVS | DJ | 55.83% | 94.17% | 84.17% | 48.33% | 57.50% |
|  | DS | 25.00% | 93.33% | 73.33% | 50.00% | 51.67% |
|  | PJ | 58.33% | 90.00% | 80.00% | 70.00% | 56.67% |
|  | PDR | 25.00% | 33.33% | 45.00% | 48.33% | 31.67% |
|  | PP | 55.00% | 71.67% | 61.67% | 68.33% | 30.00% |
|  | Total | 45.83% | 79.44% | 71.39% | 55.56% | 47.50% |
| TDS | DJ | 48.83% | 91.67% | 63.33% | 46.67% | 59.17% |
|  | DS | 31.67% | 95.00% | 78.33% | 41.67% | 45.00% |
|  | PJ | 60.00% | 95.00% | 58.33% | 65.00% | 63.33% |
|  | PDR | 31.67% | 33.33% | 51.67% | 48.33% | 31.67% |
|  | PP | 46.67% | 68.33% | 60.00% | 70.00% | 30.00% |
|  | Total | 44.45% | 79.17% | 62.50% | 53.06% | 48.06% |

## 4.2 Phase II: effect of memory representations

### 4.2.1 *Experimental setup*

Phase II examines the effect of memory representations on spatial understanding. The experimental settings in this phase are identical to those in Phase I (***Section* 4.1.1**), including the map environments, observation range, action space, navigation policy, task set, and evaluation metrics. To isolate the role of memory, we fix the exploration strategy as the Nearest-POI Strategy (NPS) and adopt a simple prompting scheme for all experiments in this phase**.** The exploring choice is motivated by the results of Phase I (***Section* 4.1.2**), which show that structured exploration strategies (NPS and TDS) yield more stable and slightly stronger performance than stochastic exploration (RVS), and NPS is also commonly adopted as a baseline strategy in local map exploration.

Four kinds of memories are compared: Simple Dialogue Memory (SDM), Node-Sequence Memory (NSM), Graph Memory (GM), and Map Memory (MM), as defined in ***Section* 3.2.3**. We evaluate their effectiveness under two settings. Specifically, we assess each memory representation in isolation, and also evaluate hybrid configurations where structured memory representations are used in conjunction with SDM.

### 4.2.2 *Result*

Table 4 reports the spatial reasoning accuracy of five foundation model agents under four explicit memory representations (SDM, NSM, GM, MM) and their hybrid variants combined with SDM (NSM+SDM, GM+SDM, MM+SDM). We analyze the results from four aspects: tasks, memory representations, models, and hybrid versus single memory.

**(1) Task-type sensitivity**

Across different memory representations, the task-level performance pattern remains highly consistent with the findings reported in Phase I (***Section* 4.1.2**) except for direction judgment (DJ). The proximity judgment (PJ) consistently emerges as the easiest tasks, POI density recognition (PDR) remains the most difficult, while distance estimation (DS) and path planning (PP) exhibit intermediate difficulty with pronounced sensitivity to memory design.

Specifically, PJ remains consistently high-performing across all memories. Even in the weakest cases, PJ does not fall below 51.67% (GPT-5.2 under SDM), while stronger models achieve substantially higher accuracy, such as Gemini-2.5-Pro maintaining PJ above 70.00% across all memory types and reaching 90.00% under NSM. In contrast, PDR is consistently the lowest-performing task. Across models and memory representations, PDR accuracy is typically below 50%, and in some cases below 20% (e.g., 18.33% for DeepSeek-V3.2 under SDM). Although structured memories can improve PDR for certain models (e.g., GPT-5.2 from 35.00% under SDM to 63.33% under NSM), overall performance remains substantially lower than for other tasks. DJ, DS, and PP—show pronounced variation across memory representations. For example, DJ for GPT-5.2 ranges from 31.67% under GM to 75.00% under NSM, and DS ranges from 20.00% under SDM to 81.67% under NSM. A similar pattern is observed for PP, which increases from 50.00% under SDM to above 90% under NSM, GM, and MM for several models. These results indicate that DJ, DS, and PP are more sensitive to changes in memory representation than PJ.

Overall, PJ remains consistently high-performing across memory representations, PDR is consistently the most challenging task, while DJ, DS, and PP exhibit substantial variation with changes in memory.

**(2) Effect of memory representations**

Comparing different memory representations, clear and systematic performance differences can be observed across tasks and models, as summarized below. NSM delivers the strongest and most consistent performance gains across tasks and models. In terms of total accuracy, NSM achieves the highest scores for all five models, ranging from 55.83% (DeepSeek-V3.2) to 86.11% (Gemini-2.5-Pro), compared with 43.61–79.44% under SDM. At the task level, NSM leads to substantial improvements in distance estimation (DS) and path planning (PP). For example, GPT-5.2's DS increases from 20.00% (SDM) to 81.67% (NSM), and PP increases from 50.00% to 91.67%. Similar gains are observed for other models, such as Claude-4.5, whose PP improves from 58.33% to 95.00%. Improvements are also evident for PDR, though performance remains lower than for PJ and PP.

GM exhibits a distinctive performance pattern characterized by very strong PP accuracy but comparatively weaker performance on DJ and DS. Under GM, PP consistently reaches high values across all models, such as 91.67–95.00% for GPT-5.2, Gemini-2.5-Pro, and Claude-4.5, and remains above 80.00% for Qwen-3 and DeepSeek-V3.2. However, DJ and DS under GM are substantially lower than under NSM for several models. For instance, GPT-5.2 achieves only 31.67% on DJ and 33.33% on DS under GM, compared with 75.00% and 81.67% under NSM. As a result, total accuracy under GM remains limited, ranging from 41.67% to 76.67%, despite the strong PP performance.

MM demonstrates a more balanced but moderate performance profile. Similar to GM, MM achieves high PP accuracy across models, such as 93.33% for GPT-5.2 and Gemini-2.5-Pro, and 88.33% for DeepSeek-V3.2. PJ under MM also remains relatively strong, typically above 70.00% for most models. In contrast, DS under MM is noticeably weaker than under NSM for several models. For example, GPT-5.2's DS under MM is 45.00%, compared with 81.67% under NSM, and Gemini-2.5-Pro's DS drops to 51.67% under MM from 96.67% under NSM. Consequently, total accuracy under MM (51.94–72.22%) is consistently lower than that under NSM, though generally higher than GM for some models.

Overall, NSM yields the strongest and most consistent performance across tasks, whereas GM and MM maintain strong path planning accuracy but show clear limitations in directional and distance-based reasoning.

**(3) Cross-model consistency**

Across all memory representations, Gemini-2.5-Pro consistently achieves the highest overall accuracy, with totals ranging from 72.22% (MM) to 86.11% (NSM), and maintains strong performance across all task categories. Claude-4.5 and Qwen-3 form a middle tier, with total accuracy typically between 57.50% and 78.06%, depending on memory type. In contrast, GPT-5.2 and DeepSeek-V3.2 show lower baseline performance but benefit markedly from structured memory. For GPT-5.2, total accuracy increases from 43.89% (SDM) to 77.78% (NSM), while for DeepSeek-V3.2 it rises from 43.61% to 55.83%. However, even with improved memory, cross-model performance gaps remain evident, indicating that memory improvements do not fully eliminate differences in model capability.

Overall, model-level differences remain evident across all memory representations, and while structured memory improves performance for weaker models, it does not eliminate cross-model performance gaps.

(4) Hybrid memory vs. single memory

As shown in Table 4, hybrid configurations generally improve performance over SDM alone, with the magnitude of gains varying by memory type and model. Among all hybrid settings, NSM+SDM yields the largest and most consistent improvements. Total accuracy increases substantially compared with SDM for all models, for example from 43.89% to 77.50% for GPT-5.2, from 79.44% to 86.67% for Gemini-2.5-Pro, and from 43.61% to 65.00% for DeepSeek-V3.2. At the task level, NSM+SDM improves DS (e.g., GPT-5.2: 20.00% → 80.00%), PP (e.g., Claude-4.5: 58.33% → 98.33%), and PDR for most models, indicating broad gains beyond those obtained with SDM alone.

GM+SDM also improves over SDM in total accuracy for most models (e.g., GPT-5.2: 43.89% → 61.11%, DeepSeek-V3.2: 43.61% → 61.67%), largely driven by strong PP performance, which remains above 80% across models. However, gains in DJ and DS are more moderate and remain lower than those achieved under NSM+SDM.

MM+SDM shows a similar but more moderate trend. While PP remains high across models (e.g., 93.33% for GPT-5.2 and Claude-4.5), improvements in DS and DJ are limited for several models, resulting in total accuracy that is generally higher than SDM but lower than NSM+SDM (e.g., GPT-5.2: 60.56%, Gemini-2.5-Pro: 77.22%).

Overall, combining structured memory with dialogue memory consistently outperforms single dialogue memory, with NSM+SDM providing the most substantial and robust gains, while GM+SDM and MM+SDM offer more limited improvements, primarily driven by path planning performance.

Table 4. The spatial reasoning performance of different foundation model agents under four memory representations.

| | Category | GPT-5.2 | Gemini-2.5-pro | Claude-4.5 | QWEN-3 | DeepSeek-V3.2 |
|---|---|---|---|---|---|---|
| SDM | DJ | 53.33% | 93.33% | 79.17% | 52.50% | 51.67% |
| | DS | 20.00% | 95.00% | 56.67% | 65.50% | 56.67% |
| | PJ | 51.67% | 91.67% | 81.67% | 73.33% | 60.00% |
| | PDR | 35.00% | 41.67% | 48.33% | 50.00% | 18.33% |
| | PP | 50.00% | 61.67% | 58.33% | 73.33% | 23.33% |
| | Total | 43.89% | 79.44% | 67.22% | 61.11% | 43.61% |
| NSM | DJ | 75.00% | 94.17% | 78.33% | 40.83% | 52.50% |
| | DS | 81.67% | 96.67% | 75.00% | 73.33% | 65.00% |
| | PJ | 80.00% | 88.33% | 90.00% | 78.33% | 76.67% |
| | PDR | 63.33% | 61.67% | 51.67% | 55.00% | 30.00% |
| | PP | 91.67% | 91.67% | 95.00% | 85.00% | 58.33% |
| | Total | 77.78% | 86.11% | 78.06% | 62.22% | 55.83% |
| GM | DJ | 31.67% | 90.00% | 51.67% | 33.33% | 25.00% |
| | DS | 33.33% | 78.33% | 56.67% | 53.33% | 28.33% |
| | PJ | 60.00% | 61.67% | 65.00% | 60.00% | 60.00% |
| | PDR | 48.33% | 45.00% | 28.33% | 51.67% | 23.33% |
| | PP | 91.67 | 95.00% | 91.67% | 80.00% | 88.33% |
| | Total | 49.44% | 76.67% | 57.50% | 51.94% | 41.67% |
| MM | DJ | 42.50% | 84.17% | 50.83% | 45.00% | 32.50% |
| | DS | 45.00% | 51.67% | 36.67% | 50.00% | 56.67% |
| | PJ | 70.00% | 70.00% | 76.67% | 81.67% | 75.00% |
|  | PDR | 61.67% | 50.00% | 38.33% | 48.33% | 26.67% |
|  | PP | 93.33% | 93.33% | 91.67% | 81.67% | 88.33% |
|  | Total | 59.16% | 72.22% | 57.50% | 58.61% | 51.94% |
| NSM+SDM | DJ | 82.50% | 94.17% | 80.00% | 45.00% | 62.50% |
|  | DS | 80.00% | 96.67% | 76.67% | 75.00% | 73.33% |
|  | PJ | 85.00% | 91.67% | 85.00% | 71.67% | 80.00% |
|  | PDR | 56.67% | 53.33% | 48.33% | 48.33% | 38.33% |
|  | PP | 78.33% | 90.00% | 98.33% | 95.00% | 73.33% |
|  | Total | 77.50% | 86.67% | 78.06% | 63.33% | 65.00% |
| GM+SDM | DJ | 53.33% | 90.83% | 75.83% | 45.00% | 61.67% |
|  | DS | 55.00% | 95.00% | 66.67% | 53.33% | 45.00% |
|  | PJ | 63.33% | 63.33% | 66.67% | 73.33% | 81.67% |
|  | PDR | 56.67% | 51.67% | 30.00% | 55.00% | 35.00% |
|  | PP | 85.00% | 91.67% | 95.00% | 80.00% | 85.00% |
|  | Total | 61.11% | 80.56% | 68.33% | 58.61% | 61.67% |
| MM+SDM | DJ | 52.50% | 90.00% | 75.83% | 46.67% | 62.50% |
|  | DS | 40.00% | 63.33% | 50.00% | 61.67% | 61.67% |
|  | PJ | 71.67% | 83.33% | 76.67% | 73.33% | 83.33% |
|  | PDR | 53.33% | 51.67% | 40.00% | 51.67% | 26.67% |
|  | PP | 93.33% | 85.00% | 93.33% | 86.67% | 86.67% |
|  | Total | 60.56% | 77.22% | 68.61% | 61.11% | 63.89% |

## 4.3 Phase III: effect of prompting and reasoning

### 4.3.1 *Experimental setup*

Phase III examines the effect of prompting and reasoning strategies on spatial understanding. Unless otherwise specified, the experimental settings in this phase are consistent with those in Phase I (***Section 4.1.1***) and Phase II (***Section 4.2.1***). To isolate the role of prompting and reasoning, we fix the exploration strategy to the Nearest-POI Strategy (NPS) and fix the memory representation to Node-Sequence Memory (NSM) for all experiments in this phase. This choice is motivated by the results of Phase II (***Section 4.2.2***), which show that NSM provides a stable improvement over simple dialogue memory. Under this controlled setting, we systematically vary the prompting and reasoning schemes, including Default Thought (DT), Chain-of-Thought (CoT), Self-Consistency with CoT (SC-CoT), and Tree-of-Thoughts (ToT), as defined in ***Section 3.3.1***.

### 4.3.2 *Result*

Table 5 reports the spatial reasoning performance of five foundation model agents under DT, CoT, SC-CoT, and ToT while fixing the exploration strategy to NPS and the memory representation to NSM. Following the analysis structure adopted in ***Sections 4.1.2 and 4.2.2***, we also analyze the results from three perspectives: asks, exploration strategies, and models.

(1) Task-type sensitivity

Across all prompting schemes and models, a stable hierarchy among task categories is again observed, consistent with the patterns reported in *Sections* **4.1.2 and 4.2.2**. Direction judgment (DJ) and proximity judgment (PJ) remain the most reliable tasks, while POI density recognition (PDR) remains the most challenging. DJ exhibits strong and consistent performance across reasoning schemes. For example, Gemini-2.5-Pro maintains DJ accuracy above 92.5% under all prompts, while Claude-4.5 improves DJ from 78.33% (DT) to 91.67% under ToT. Even weaker models benefit from advanced prompting: GPT-5.2 improves from 75.00% (DT) to 80.83% (ToT), and DeepSeek-V3.2 rises from 52.50% (DT) to 72.50% (T). These results indicate that directional reasoning can be reliably enhanced by explicit reasoning guidance once sufficient sequential memory is available. PJ also remains consistently high. Across models, PJ accuracy typically exceeds 75% under most prompting schemes and reaches particularly high values under SC-CoT and ToT (e.g., DeepSeek-V3.2 achieves 90.00% under both CoT-based schemes). This stability echoes earlier findings that proximity judgments rely primarily on local relational information and are less sensitive to both exploration and memory changes.

In contrast, POI density recognition (PDR) continues to exhibit the lowest overall accuracy across reasoning schemes. While reasoning prompts do improve PDR for some models—most notably Qwen-3, which achieves 80.00% under SC-CoT compared to 55.00% under DT—performance remains unstable and model-dependent. For Gemini-2.5-Pro, PDR remains around 50–61% across all schemes, indicating that even strong reasoning prompts cannot fully compensate for the intrinsic difficulty of global density estimation under partial observability. This pattern is consistent with Phase I and Phase II, where PDR also showed limited sensitivity to strategy and memory changes.

Distance estimation (DS) and path planning (PP) exhibit the most pronounced sensitivity to prompting strategies. For example, Claude-4.5 improves DS from 75.00% (DT) to 90.00% (ToT), and PP from 95.00% (DT) to 91.67%–96.67% across advanced schemes. Similar trends are observed for Qwen-3 and DeepSeek-V3.2, indicating that explicit reasoning structures are particularly beneficial for tasks requiring multi-step spatial integration.

Overall, the task-level hierarchy observed in earlier phases remains stable: PJ and DJ are consistently high-performing, PDR remains challenging, while DS and PP are the most responsive to prompting and reasoning enhancements.

**(2) Effect of prompting and reasoning schemes**

Comparing prompting strategies, ToT yields the strongest overall performance gains across models and tasks. For all five models, ToT achieves the highest total accuracy, improving upon DT by 3-18 percentage points. For example, GPT-5.2 increases from 77.78% (DT) to 81.39% (ToT), Claude-4.5 from 78.06% to 84.17%, and DeepSeek-V3.2 from 55.83% to 73.89%.

SC-CoT also provides consistent improvements over DT and standard CoT. In particular, SC-CoT leads to substantial gains in PP and DS for several models, such as Qwen-3, whose total accuracy increases from 62.22% (DT) to 73.61% under SC-CoT. These results suggest that sampling multiple reasoning paths helps stabilize reasoning outcomes for spatial tasks involving uncertainty and cumulative inference.

Standard CoT produces mixed effects. While it improves certain tasks for specific models—such as PP for DeepSeek-V3.2 (58.33% → 88.33%)—it does not consistently outperform DT in total accuracy across all models. This indicates that simply encouraging step-by-step reasoning is not always sufficient and may introduce noise without aggregation or selection mechanisms.

Overall, these results indicate that advanced reasoning schemes that explicitly structure or select reasoning paths (SC-CoT and ToT) are more effective than shallow prompting in improving spatial reasoning, particularly for tasks requiring integration of multiple spatial cues.

**(3) Cross-model consistency**

Across all prompting schemes, cross-model performance differences remain evident, consistent with earlier phases. Gemini-2.5-Pro consistently achieves the highest total accuracy, with performance remaining above 84% under all reasoning schemes. Claude-4.5 follows closely, particularly under SC-CoT and ToT. Qwen-3 and DeepSeek-V3.2 benefit substantially from advanced prompting but still lag behind stronger models, indicating that prompting cannot fully compensate for baseline model capability.

Importantly, the relative ranking of models remains stable across reasoning schemes, mirroring the cross-model consistency observed in ***Sections* 4.1.2 and 4.2.2**. This suggests that while prompting and reasoning schemes can significantly enhance performance, they primarily amplify existing spatial capabilities rather than reshuffle model-level strengths.

Table 5. The spatial reasoning performance of different foundation model agents under four reasoning schemes-Default Memory (DT), Chain-of-thought (CoT), Self-consistency CoT (SC-CoT), and Tree-of-thought (ToT).

| | Category | gpt-5.2 | gemini-2.5-pro | claude-4.5 | qwen-3 | DeepSeek-V3.2 |
|---|---|---|---|---|---|---|
| | DJ | 75.00% | 94.17% | 78.33% | 40.83% | 52.50% |
| DT | DS | 81.67% | 96.67% | 75.00% | 73.33% | 65.00% |
| | PJ | 80.00% | 88.33% | 90.00% | 78.33% | 76.67% |
|  | PDR | 63.33% | 61.67% | 51.67% | 55.00% | 30.00% |
|  | PP | 91.67% | 91.67% | 95.00% | 85.00% | 58.33% |
|  | Total | 77.78% | 86.11% | 78.06% | 62.22% | 55.83% |
| CoT | DJ | 75.83% | 94.17% | 87.50% | 79.17% | 52.50% |
|  | DS | 81.67% | 96.67% | 81.67% | 60.00% | 65.00% |
|  | PJ | 83.33% | 88.33% | 75.00% | 58.33% | 90.00% |
|  | PDR | 58.33% | 50.00% | 46.67% | 55.00% | 36.67% |
|  | PP | 85.00% | 88.33% | 96.67% | 73.33% | 88.33% |
|  | Total | 76.67% | 85.28% | 79.17% | 67.50% | 64.17% |
| SC-CoT | DJ | 75.83% | 93.33% | 90.00% | 74.17% | 54.17% |
|  | DS | 81.67% | 95.00% | 88.33% | 65.00% | 66.67% |
|  | PJ | 86.67% | 90.00% | 76.67% | 65.00% | 90.00% |
|  | PDR | 58.33% | 51.67% | 51.67% | 80.00% | 40.00% |
|  | PP | 90.00% | 93.33% | 93.33% | 83.33% | 90.00% |
|  | Total | 78.06% | 86.11% | 81.67% | 73.61% | 65.83% |
| ToT | DJ | 80.83% | 92.50% | 91.67% | 71.67% | 72.50% |
|  | DS | 83.33% | 93.33% | 90.00% | 78.33% | 76.67% |
|  | PJ | 80.00% | 86.67% | 86.67% | 90.00% | 88.33% |
|  | PDR | 68.33% | 50.00% | 53.33% | 45.00% | 43.33% |
|  | PP | 95.00% | 90.00% | 91.67% | 76.67% | 90.00% |
|  | Total | 81.39% | 84.17% | 84.17% | 72.22% | 73.89% |

## 5. Discussion

### 5.1 Impact of different strategies

#### 5.1.1 *How agent explore as experience acquisition in a map environment?*

Exploration plays a fundamental role in how foundation model (FM) agents acquire spatial experience in symbolic map environments. Results from Phase I (***Section* 4.1**) show that varying exploration strategies leads to only modest differences in overall spatial reasoning accuracy. Building on these findings, we further examine and try to explain how such differences manifest across tasks and models by visualizing performance at both the task level and the model level, see Figure 5.

As illustrated in Figure 5 (left), task-level performance patterns remain highly consistent across different exploration strategies. Direction judgment (DJ) and proximity judgment (PJ) consistently achieve higher accuracy across all strategies, whereas POI density recognition (PDR) remains the most challenging task. Even under the weakest configurations, PJ accuracy stays above 50%, while PDR frequently falls below 40% across models and strategies. This stability in task difficulty ordering suggests that exploration strategy alone does not fundamentally alter the cognitive demands associated with different spatial tasks. Instead, it indicates that certain spatial competencies—such as relative direction and local proximity—can be robustly acquired through localized observations, whereas tasks requiring global aggregation, such as density estimation, remain intrinsically difficult under partial observability. At the same time, Figure 5 (left) reveals that exploration strategy does modulate performance for specific task types. Random-Visible Strategy (RVS) yields modest but consistent improvements in path planning (PP) for several models, reflecting the benefit of stochastic exposure to diverse local routes and intersections. By contrast, Task-Driven Strategy (TDS) occasionally benefits proximity judgment, likely because repeated traversal between fixed POI pairs reinforces local relational cues. These effects, however, remain task-specific and do not substantially reshape overall performance profiles. Figure 5 (right) shows that the impact of exploration strategy on overall accuracy is limited when compared with model-level differences. For most models, total accuracy varies within a narrow range—typically most are less than 3 percentage points—across NPS, RVS, and TDS. Strong models such as Gemini-2.5-Pro remain highly stable across strategies, whereas weaker models exhibit slightly higher sensitivity but still remain constrained by their underlying spatial abstraction capabilities. This cross-model consistency indicates that exploration strategy influences how experiences are sampled, but not how effectively they are integrated.

Taken together, we can see that exploration primarily affects the distribution and ordering of experiential evidence, rather than the quality of spatial understanding itself. The reason may be that: under the goal-driven interaction framework with shortest-path navigation, all exploration strategies eventually ensure sufficient spatial coverage and repeated POI encounters. As a result, strategy-level differences manifest mainly in local relational tasks, while higher-order tasks requiring cumulative spatial integration—such as global density or distance aggregation—remain largely unaffected. These findings suggest that exploration is a necessary but insufficient condition for spatial understanding: it supplies raw experience, but meaningful improvements in global spatial reasoning depend critically on how experience is consolidated through memory representations, as examined in Phase II.

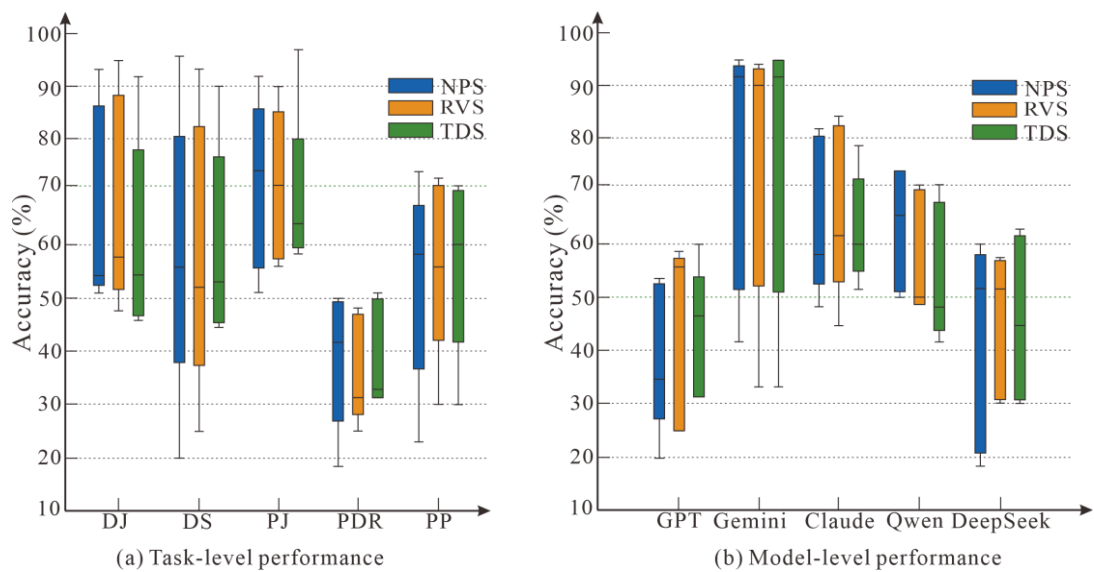

Figure 5. The task-level and model level performance of different foundation model agents under three exploration strategies (NPS, RVS and TDS).

**5.1.2 *How agent memory as experience consolidation in a map environment?***

While exploration determines how spatial experience is acquired, memory plays a decisive role in how such experience is consolidated, structured, and subsequently exploited for reasoning. In symbolic map environments, where observations are partial and sequential, memory mechanisms determine whether accumulated local evidence can be transformed into coherent spatial knowledge. Results in ***Section* 4.2.2** show that changing memory representations induces substantially larger performance variations than changing exploration strategies, confirming memory as a central factor in spatial reasoning performance. To clarify and explain how different memory designs contribute to these effects, Figure 6 summarizes the key results by contrasting single structured memories and hybrid memory configurations relative to Simple Dialogue Memory (SDM).

The left panel of Figure 6 compares three structured memory representations—Node-Sequence Memory (NSM), Graph Memory (GM), and Map Memory (MM)—each used alone, against SDM. From this panel, it can be observed that structured memories reshape task-level performance in markedly different ways. Proximity judgment (PJ) remains relatively robust across memory types, with NSM yielding a moderate improvement of about +11%, whereas GM incurs a clear decline. This pattern can be explained by the local nature of proximity reasoning. Sequential memory preserves neighborhood-level encounter order, which supports relative comparison, whereas graph abstraction may suppress such fine-grained local relations by prioritizing global connectivity. POI density recognition (PDR) remains challenging across all memory representations, with only limited gains from structured memory. This outcome can be attributed to the global and aggregative nature of density reasoning, which requires integrating spatial evidence across the entire map. Under partial observability, even well-structured memories struggle to compensate for incomplete coverage, leading to persistently low performance regardless of memory type.

More pronounced memory effects appear in direction judgment (DJ), distance estimation (DS), and path planning (PP). Notably, NSM produces consistent positive gains across these tasks, including approximately +20% in DS and over +30% in PP, resulting in an overall improvement of nearly +13% relative to SDM. This pattern can be explained by NSM's ability to preserve the temporal order of visited locations and traversed paths, enabling cumulative spatial integration over time. In contrast, GM and MM show strong gains in PP (around +36%), but suffer substantial losses in DJ and DS, leading to neutral or even negative overall effects. This trade-off reflects the emphasis of graph- and map-based representations on global structure and connectivity, which facilitates route planning, while reducing access to sequential and egocentric cues that are critical for directional and metric reasoning.

The right panel of Figure 6 illustrates the effect of hybrid memory configurations, in which each structured memory is combined with SDM and compared against SDM alone. From this panel, it is evident that hybrid memory consistently improves overall performance across models. In particular, NSM+SDM achieves the strongest and most balanced gains, exceeding +20% in DS, approximately +34% in PP, and yielding an overall improvement of over +15%. These improvements can be explained by the complementary roles of the two memory types: dialogue memory preserves recent observations and linguistic context, while structured memory provides an explicit organization of spatial experience. Together, they enable more flexible access to both local and global spatial cues during reasoning. Furthermore, GM+SDM and MM+SDM also achieve clear overall improvements (around +7%), primarily driven by large PP gains while mitigating the sharp declines in DJ and DS observed when these memories are used alone. This indicates that dialogue memory can partially offset the abstraction effects introduced by graph- or map-based representations, preserving fine-grained spatial cues that would otherwise be attenuated.

Taken together, we can conclude that single structured memories already alter how spatial experience is integrated, but their effectiveness depends critically on how well the memory structure aligns with the nature of the spatial reasoning task. Hybrid memory configurations further enhance robustness by combining explicit spatial organization with flexible dialogue context. Sequential memory representations that preserve traversal order are particularly effective for tasks requiring cumulative spatial integration, whereas more abstract memory forms provide selective benefits concentrated on structural reasoning. These findings highlight that effective spatial understanding depends not merely on storing more information, but on how experiential evidence is organized and integrated within memory, establishing memory design as a central bottleneck—and opportunity—for foundation model agents operating in map environments.

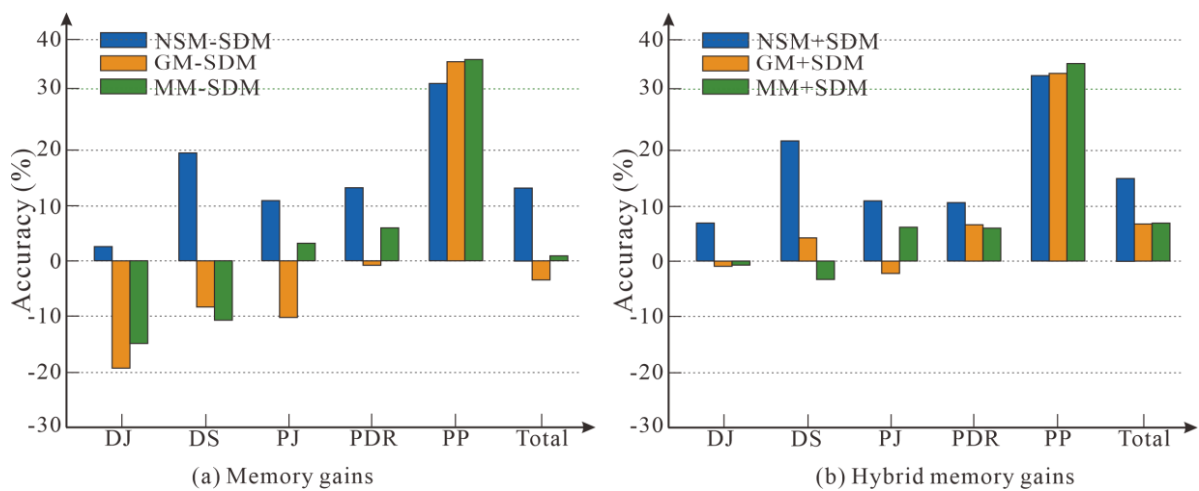

Figure 6. Memory gains by task. Left (a): Task-level gains of structured memory representations relative to SDM. Right (b): Overall gains of hybrid memory configurations relative to SDM. Results are averaged across five foundation models.

### 5.1.3 *How agent reason as experience utilization in a map environment?*

While exploration governs how spatial experience is acquired and memory determines how experience is consolidated, prompting and reasoning mechanisms primarily influence how consolidated experience is utilized during inference. Results from Phase III (***Section*** **4.3.2**) show that prompting and reasoning schemes primarily modulate the expression of spatial reasoning capability, rather than altering the fundamental structure of the spatial tasks themselves. Building on these findings, we further examine and try to explain how such differences manifest across tasks and models by visualizing the accuracy improvement from Default Thought (DT) to Tree-of-Thoughts (ToT), see Figure 7.

As further illustrated in Figure 7, the impact of prompting manifests primarily as capacity-dependent performance gains rather than task reordering. Specifically, models with lower baseline accuracy benefit substantially more from increased prompt complexity, whereas high-capacity models exhibit only marginal gains. For example, DeepSeek-V3.2 and Qwen-3 show pronounced improvements exceeding 10 percentage points, while Gemini-2.5-Pro remains nearly flat despite increasingly sophisticated reasoning prompts. Examining task-level improvements for weaker models further indicates that these gains are concentrated in tasks requiring multi-step spatial integration, particularly distance estimation (DS) and path planning (PP). Concretely, as shown in Table 4, DeepSeek-V3.2 improves from 55.83% under DT to 73.89% under ToT, while Qwen increases from 62.22% to 72.22% over the same prompting progression. These patterns confirm that prompting mainly enhances the effective utilization of existing spatial evidence and multi-step reasoning processes, particularly for models whose spatial abstraction and inference stability are initially limited.

Taken together, we can see that the prompting is most effective as a capability amplifier for weaker models, improving the utilization of available spatial evidence and stabilizing multi-step reasoning. However, for stronger models, prompt-induced structure offers limited additional benefit and does not overcome task-intrinsic difficulty. This asymmetry reinforces the view that prompting effectiveness is fundamentally constrained by both the nature of the task and the baseline spatial competence of the model.

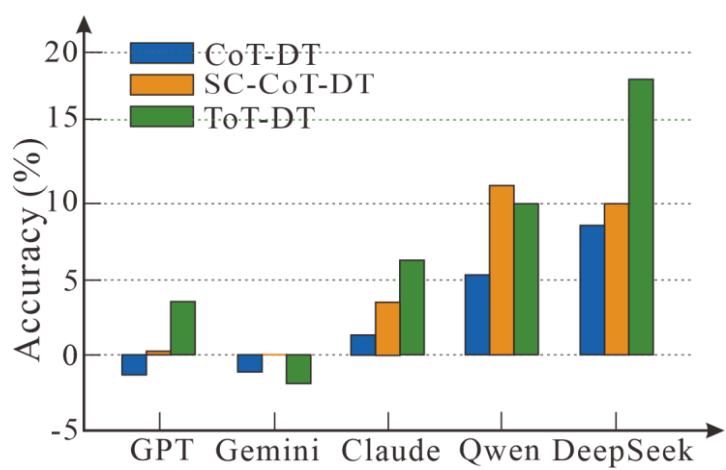

Figure 7. The overall accuracy gain from Default Thought to Tree-of-Thoughts across models.

### 5.1.4 *Design implication for FM map Agents*

Based on all findings across Phases I–III, several implications for the design of foundation model (FM) agents operating in symbolic map environments can be drawn. Rather than treating exploration, memory, and prompting as interchangeable optimization levers, our results suggest that these components play distinct and complementary roles in shaping spatial understanding, and should therefore be designed and coordinated with different objectives in mind.

**(1) Exploration strategies should be designed to ensure coverage rather than to optimize task performance directly**. Phase I shows that while exploration strategy influences the distribution of local experiences, its overall effect on spatial reasoning accuracy is bounded. Task difficulty hierarchies remain largely stable across different

exploration strategies, and total performance varies within a narrow range for most models. This suggests that, beyond guaranteeing sufficient spatial exposure, increasingly sophisticated exploration policies yield diminishing returns unless paired with effective mechanisms for consolidating experience. In practice, this implies that simple, structured exploration strategies—such as nearest-neighbor–based exploration—are often sufficient as a foundation, allowing system complexity to be focused elsewhere.

**(2) Memory representation should be treated as the primary design axis for improving spatial reasoning capability**. Phase II demonstrates that different memory structures fundamentally reshape how accumulated experience is integrated and exploited. Sequential memory representations, such as Node-Sequence Memory, consistently yield the strongest and most stable improvements across tasks, particularly for those requiring cumulative spatial integration. More abstract representations, including graph- or map-based memories, provide selective benefits for structural reasoning tasks (e.g., path planning) but may attenuate fine-grained directional or metric information. These findings imply that memory design should be task-aware and that hybrid memory architectures—combining structured spatial representations with flexible dialogue memory—offer a robust and generalizable solution for FM map agents.

**(3) Prompting and reasoning mechanisms should be viewed as adaptive capability amplifiers rather than universal performance boosters**. Phase III reveals that the effectiveness of prompting is strongly conditioned on baseline model capacity. Structured prompting schemes substantially benefit weaker models by stabilizing inference and supporting multi-step reasoning, while stronger models exhibit diminishing returns. Importantly, prompting does not alter task-intrinsic difficulty nor compensate for missing spatial evidence; instead, it governs how effectively existing knowledge is utilized. From a system design perspective, this suggests that prompt complexity should be adjusted dynamically based on model capacity and task demands, rather than uniformly increased.

## 5.2 Trade-offs between memory size and spatial reasoning performance

Beyond absolute task accuracy, an important but often overlooked aspect of experience-based spatial reasoning lies in the trade-off between memory size, measured explicitly in bits, and reasoning effectiveness. In Table 6, we reported the memory size of each memory representation as the total number of bits.

A clear and consistent pattern emerges across all evaluated foundation models. Simple Dialogue Memory (SDM) exhibits by far the largest memory size, exceeding 20,000 bits for all models. This high cost arises because SDM relies on raw dialogue context, in which observations, actions, and intermediate descriptions are stored as unstructured token sequences, leading to substantial redundancy. Despite this large memory footprint, SDM yields the lowest overall spatial reasoning accuracy. For example, GPT-5.2 consumes over 20,800 bits under SDM yet achieves only 43.89% accuracy, indicating that increasing memory size alone does not guarantee effective spatial understanding.

In contrast, Node-Sequence Memory (NSM) achieves a substantially better balance between memory size and reasoning performance. Across all models, NSM reduces memory usage by approximately 45–50% relative to SDM, with memory sizes concentrated around 11,115.3-11,519.3 bits, while simultaneously delivering the highest overall accuracy. For instance, GPT-5.2 improves from 43.89% (SDM) to 77.78% (NSM) while reducing memory size from 20,870.9 bits to 11,115.3 bits. This improvement demonstrates that explicitly encoding experience as ordered node sequences removes redundant contextual information while preserving essential spatial cues, resulting in a more memory-efficient and reasoning-effective representation.

More compact representations, such as Graph Memory (GM) and Map Memory (MM), further compress memory size to the range of 2,730.8-3,122.6 bits and 3,704.3-3,844.6 bits, respectively. While this aggressive compression greatly reduces memory cost, it also introduces a performance trade-off. Although GM and MM retain strong performance on structurally oriented tasks such as path planning, their overall accuracy consistently falls below that of NSM. For example, GPT-5.2 reaches only 49.44% under GM and 59.16% under MM, despite their much smaller memory footprints. Similar trends are observed across all evaluated models. These results suggest that highly abstract memories, while efficient in terms of bits, may discard fine-grained sequential or metric information necessary for tasks such as distance estimation and directional reasoning.

Taken together, these findings reveal a non-linear relationship between memory size (in bits) and spatial reasoning performance. Larger memory does not necessarily yield better reasoning, and extreme compression can impair task performance by eliminating task-relevant experiential detail. Instead, there exists an intermediate 'sweet spot' in which memory representations are sufficiently structured to reduce redundancy while still rich enough to preserve sequential and local spatial information. In our experiments, Node-Sequence Memory consistently occupies this optimal region, offering the most favorable accuracy–bit trade-off across all models.

Table 6. Memory size and overall spatial reasoning accuracy under different memory representations.

| Category | | GPT-5.2 | Gemini-2.5-pro | Claude-4.5 | Qwen-3 | DeepSeek-V3.2 |
|---|---|---|---|---|---|---|
| Memory size | SDM | 20870.9 | 21157.9 | 25372.9 | 22731.7 | 25417.4 |
| | NSM | 11115.3 | 11411.9 | 11287.6 | 11169.5 | 11519.3 |
| | GM | 2730.8 | 3063.1 | 2912.6 | 3005.0 | 3122.6 |
| | MM | 3704.3 | 3801.4 | 3799.7 | 3764.7 | 3844.6 |
| Total accuracy | SDM | 43.89% | 79.44% | 67.22% | 61.11% | 43.61% |
| | NSM | 77.78% | 86.11% | 78.06% | 62.22% | 55.83% |
| | GM | 49.44% | 76.67% | 57.50% | 51.94% | 41.67% |
| | MM | 59.16% | 72.22% | 57.50% | 58.61% | 51.94% |

## 5.3 Performance across model version and parameter size

Beyond memory representation and reasoning schemes, model capability itself is a potential confounding factor in spatial reasoning evaluation. Recent foundation models vary substantially across versions and parameter scales, raising an important question: to what extent do improvements in spatial reasoning stem from model capacity itself? To this end, we evaluate Claude (with multiple versions) and Qwen (with multiple parameter sizes) across versions and parameter sizes, while keeping the task settings, exploration strategy (NPS), memory representations (NSM), and reason scheme (DT) fixed. The results are shown in Figure 8.

Figure 8a (left) illustrates the impact of model version on spatial reasoning accuracy. As shown, performance remains largely stable across successive model versions, with only minor fluctuations in overall accuracy. In particular, models based on Claude already exhibit strong and consistent spatial reasoning performance. It is worth noting that Claude 3.7, released earlier in the model lineage, already demonstrates highly competitive results across spatial tasks, suggesting that its underlying spatial capabilities are close to saturation. Subsequent version updates do not lead to substantial accuracy gains, indicating that improvements in general language modeling or reasoning proficiency do not necessarily translate into further improvements in basic spatial understanding. Figure 8b (right) shows the effect of model parameter size on spatial reasoning accuracy. Accuracy increases rapidly when moving from smaller to medium-sized models, reflecting the benefits of increased representational and reasoning capacity at early stages of scaling. However, once model size reaches a sufficiently large regime, performance improvements plateau. Beyond this point, further increases in model size yield only marginal or negligible gains across most spatial tasks.

Taken together, these two observations suggest that basic spatial reasoning abilities in foundation models reach a bottleneck once a certain capability level is achieved. While early improvements can be attributed to increased model capacity, further scaling—either through newer versions or larger parameter counts—does not substantially enhance spatial performance. This implies that limitations in map-based spatial reasoning are not primarily due to insufficient model size or general reasoning power. Instead, these findings indicate that advancing spatial cognition in FM agents requires mechanisms beyond model scaling, such as explicitly structured spatial memory, experience organization, and spatially grounded reasoning processes. Once models are sufficiently strong, improvements in spatial understanding depend more on how spatial experience is represented and utilized than on how large or how recent the model is.

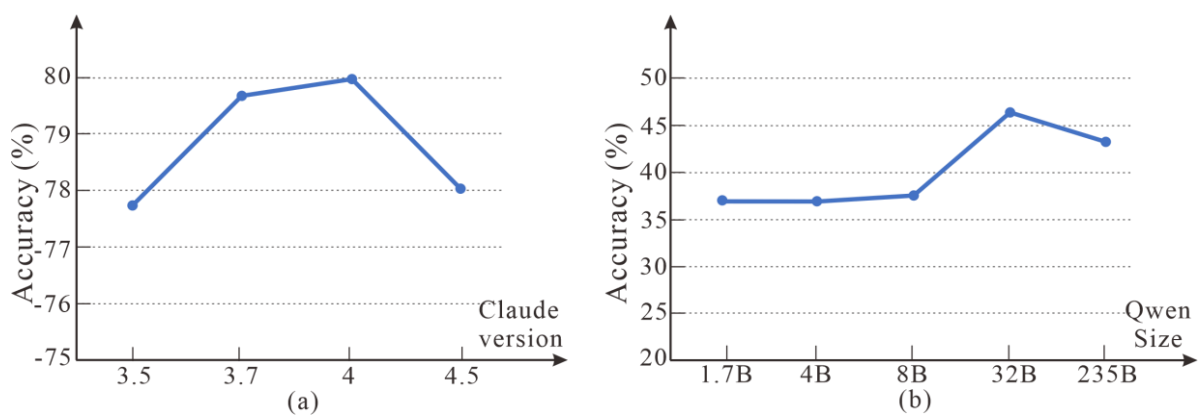

Figure 8. Effect of model version (a) and model parameter size (b) on overall spatial reasoning accuracy.

## 5.4 Why the FM agents fail and how memory and reasoning fix it

While *Sections* 4.1–4.3 quantitatively demonstrate that structured memory and reasoning schemes substantially improve spatial reasoning accuracy, these results alone do not explain how such improvements emerge. To better understand the underlying mechanisms, we analyze representative failure and success cases by examining the agents' internal reasoning traces during inference.

**(1) Simple dialogue memory vs. graph memory**

Results in *Section* 4.2.2 demonstrate that graph memory consistently improves spatial reasoning over simple dialogue memory, with the largest gains observed in structure-intensive tasks, especially the path planning task. We have selected a representative navigation case in Paris 'asks for the shortest route between Café Le Dôme and Café de l'Alma' and compare the reasoning processes of a Claude model under the two memory settings.

Under simple dialogue memory, we can observe that the model selects an incorrect path and relies on heuristic reasoning rather than explicit route reconstruction. The reasoning trace shows that the decision is based on qualitative assumptions such as a 'nearest POI strategy,' while explicitly noting that 'no concrete distances are provided,' and admitting that the choice is made 'with insufficient evidence.' This indicates that spatial experience remains fragmented and cannot be composed into a verifiable end-to-end path. In contrast, with graph memory, the model reconstructs an explicit route connecting Café Le Dôme to Café de l'Alma by enumerating ordered intermediate nodes. The decision is further justified by distance evidence (e.g., 'approximately 75 meters') and by comparing alternative paths based on their structural complexity. This shift from heuristic guessing to graph-constrained reconstruction explains why graph memory effectively repairs path planning failures.

**(2) Default thought vs. tree-of-thought**

Results in *Section* **4.3** show that reasoning schemes further modulate spatial reasoning performance once memory is fixed, with Tree-of-Thought (ToT) consistently outperforming Default Thought (DT) on multi-step tasks such as path planning and distance estimation. To illustrate this effect, we analyze a representative navigation case answered by the DeepSeek model under the two prompting strategies, using the same underlying memory.

Under DT, the reasoning trace suggests a pattern of premature commitment. The model justifies its choice with heuristic language indicating that it is selecting a single plausible route (e.g., '[DT-QUOTE-1]') while explicitly lacking systematic comparison (e.g., '[DT-QUOTE-2]'). This implies that, although the required spatial evidence is available, the inference process collapses early into one hypothesis rather than evaluating alternatives.

In contrast, under ToT prompting, the reasoning trace exhibits explicit enumeration and comparison. The model proposes multiple candidate routes (e.g., '[ToT-QUOTE-1]'), evaluates them using structural or cost-based criteria (e.g., '[ToT-QUOTE-2]'), and then selects the best-supported option (e.g., '[ToT-QUOTE-3]'). This shift from single-solution commitment to structured comparison explains why ToT more effectively activates stored spatial knowledge and yields more reliable performance on tasks requiring multi-step spatial integration.

## 5.5 Limitation and future works

By systematically evaluating how exploration strategies, memory representations, and prompting schemes influence spatial understanding in symbolic map environments, this study provides a structured view of how experience-based spatial reasoning emerges in foundation model agents. However, several limitations still should be acknowledged.

First, the proposed map environments are symbolic and discretized, relying on grid-based representations of roads and POIs. While this abstraction enables controlled comparison across cities and tasks, it inevitably simplifies geometric detail, scale variation, and cartographic conventions present in real-world maps. As a result, the evaluated spatial understanding reflects reasoning over simplified symbolic layouts rather than full-fidelity cartographic products. Future work could extend the framework to multi-scale or vector-based map representations, allowing agents to reason over more realistic spatial geometries and cartographic encodings.

Second, the exploration process assumes shortest-path navigation between selected POIs, thereby decoupling high-level exploration decisions from low-level route planning. This design choice isolates the effect of exploration strategies on experience acquisition but limits the evaluation of navigation-policy learning and adaptive movement behavior. Incorporating learned or stochastic routing policies would allow future studies to jointly examine exploration, navigation, and spatial reasoning under greater behavioral variability.

Third, although the proposed memory representations capture different forms of spatial knowledge, they remain explicitly engineered structures rather than learned or adaptive memory systems. Consequently, the evaluation focuses on how existing foundation models utilize structured memory inputs, rather than how such memories could be autonomously formed or refined. Future research may explore learned memory induction, memory compression, or continual memory updating, enabling agents to adapt their internal representations dynamically as exploration progresses.

Finally, the current evaluation focuses on task accuracy as the primary metric of spatial understanding. While accuracy provides a clear and comparable measure, it does not fully capture reasoning efficiency, robustness, or error patterns. Future work could incorporate additional metrics—such as reasoning consistency, memory efficiency, or step-level error analysis—to provide a more comprehensive assessment of spatial cognition in foundation model agents.

## 6. Conclusion

This paper investigates how foundation model (FM) agents explore, remember, and reason about symbolic map environments using an interactive evaluation framework. By shifting from static map queries to incremental exploration under partial observability, we disentangle the functional roles of exploration, memory, and reasoning in map-based spatial cognition. Our results show that exploration primarily affects how spatial experience is acquired but has limited influence on final reasoning accuracy once sufficient coverage is achieved. In contrast, memory representation plays a central role in consolidating experience into usable spatial knowledge. Structured memories, particularly sequential and graph-based representations, consistently outperform simple dialogue memory on structure-intensive tasks such as distance estimation and path planning, while hybrid memory configurations further improve robustness by combining explicit spatial organization with flexible dialogue context. Reasoning schemes further influence how stored spatial knowledge is utilized, with advanced prompts encouraging more systematic multi-step inference and reducing heuristic guessing, especially for models with weaker baseline spatial capabilities, whereas gains for stronger models are more limited. This capacity-dependent effect aligns with our observation that spatial reasoning performance tends to saturate across model versions and scales once a certain capability threshold is reached. These suggest that simply increasing model size or upgrading model versions alone is insufficient for improving map-based spatial abilities, and that further progress requires mechanisms that better structure and integrate spatial experience, particularly at the level of memory and spatial reasoning. Together, these findings clarify the complementary roles of exploration, memory, and reasoning in spatial cognition of FM agents and provide practical guidance for designing memory- and reasoning-aware map-based agents. A key limitation of this study is its focus on symbolic and planar map environments; future work will extend this framework to richer map representations and longer-horizon spatial reasoning settings.

## Acknowledgments

## Disclosure statement

## Funding

This work was supported by grants from the National Natural Science Foundation of China (No. 42501551, 42371455), Tobii China Innovation Initiative Project (TPI250407CN).

## Notes on contributors